# Improve the Autonomy of the $SE_2(3)$ group based Extended Kalman Filter for Integrated Navigation: Application

**MAOSONG WANG**

National University of Defense Technology, Changsha, Hunan 410073, China

**JIARUI CUI**

National University of Defense Technology, Changsha, Hunan 410073, China
Beijing Institute of Tracking and Communication Technology, Beijing 100094, China

**WENQI WU**

**PEIQI LI**

**XIANFEI PAN**

National University of Defense Technology, Changsha, Hunan 410073, China

Maosong Wang, Wenqi Wu, Peiqi Li and Xianfei Pan are with National University of Defense Technology, Changsha, Hunan 410073, China. (e-mail: wangmaosong12@nudt.edu.cn; wenqiwu_lit@hotmail.com; lpq1535623360@163.com; afeipan@126.com; ).

Jiarui Cui was with National University of Defense Technology, Changsha, Hunan 410073, China. He is now with the Beijing Institute of Tracking and Communication Technology, Beijing 100094, China. (e-mail: cuijr1996@alumni.nudt.edu.cn).

*Abstract*—One of the core advantages of $SE_2(3)$ Lie group framework for navigation modeling lies in the autonomy of error propagation. In the previous paper, the theoretical analysis of autonomy property of navigation model in inertial, earth and world frames was given. A construction method for $SE_2(3)$ group navigation model is proposed to improve the non-inertial navigation model toward full autonomy. This paper serves as a counterpart to previous paper and conducts the real-world strapdown inertial navigation system (SINS)/odometer(ODO) experiments as well as Monte-Carlo simulations to demonstrate the performance of improved $SE_2(3)$ group based high-precision navigation models.

## I. INTRODUCTION

Work on state estimation of navigation problem dates back to the 1960s with the introduction of the extended Kalman filter (EKF) [1][7] and it was successfully applied to integrated navigation state estimation for the Apollo missions[3]. Extended Kalman filter has essentially become the standard algorithm in industry over the past six decade [4][5][6].

Original navigation state estimation problem mainly concerns attitude estimation. The EKF is used to linearize the nonlinear attitude estimation problem on the group $SO(3)$, such as quaternions, direction cosine matrices (DCM) and Euler angles (they are used as tools to represent the 3-dimensional rotation special orthogonal group $SO(3)$ [7][8]). In EKF based navigation model, the velocity and position parameter are augmented directly to the nonlinear attitude parameter (equals to direct product group $SO(3) \times R^6$). Whereas, with the understanding of the geometry nature of rigid body dynamic in 3-demission space, the navigation states are modelled on more elegant groups, such as special Euclidean group $SE(3)$ (equals to semi-direct product group $SO(3) \ltimes R^3$) considering the nonlinear effects of attitude errors on position [9]. The $SE(3)$ has widely applied to solve pose estimation problem in the industrial and consumer-grade ground robotics[11], arial vehicles [10] and underwater vehicles[12]. The state estimation on group $SE(3)$ is naturally a nonlinear problem, thus numerous nonlinear filtering methods, such as EKF, unscented Kalman filter (UKF) [13] or particle filter (PF) [14], to simulate the nonlinear state propagation process, which have achieved better performance on position and attitude estimation to some extent. Barrua defined a novel group namely augmented special Euclidean group $SE_2(3)$ (equals to semi-direct product group $SO(3) \ltimes R^6$) that further includes velocity state and identified a group affine system based on $SE_2(3)$ [15]. This group affine system has been proven to have the linear propagation property (or autonomy property, thanks to the log linear property) of the nonlinear states in navigation problem. This led to the invariant Kalman filter (IEKF) [16] and the TFG-invariant Kalman filter (TFG-IEKF) [17] which proved to have error propagation autonomy property for state estimation problem in some classical navigation applications. Wang and Cui reformed the high-precision traditional integrated navigation formulations using matrix Lie group $SE_2(3)$ and proposed Lie group extended Kalman filter (LG-EKF [18]-[20]), while Chang also conducted related work [21][21]. Mahong also proposed a methodology of equivariant filter (EqF) [23][24], which provides a general design method for systems evolving in Lie group with biased input measurement and it could boil down to the invariant Kalman filter in many navigation applications practically [25]. The research on state estimation and integrated navigation on the $SE_2(3)$ group is one of the current cutting-edge topics.

The critical advantage of the Lie group $SE_2(3)$ based extended Kalman filter largely lies in the autonomy of error propagation. Autonomy is a natural property for linear systems (translational invariance), but not always true for nonlinear systems since attitude existent. By using the $SE_2(3)$ matrix group representation, the navigation state error could be represented by a compact nonlinear form [18]-[20]. This nonlinear error dynamic possesses a favorable log linear property by which arbitrary amount navigation error could be calculated exactly though nonlinear exponential map [15]. Additionally, the propagation of state error turns to be trajectory independent that there is no state-dependent term in the differential equation of nonlinear error. This avoids the selection of the specific linearization point, and this should be the key basis for the performance of the geometry filters (IEKF, LG-EKF, EqF) to outperform the EKF using traditional $SO(3) \times R^6$ group. However, this autonomy only holds under simplified conditions. Firstly, it must ignore sensor bias since the bias is not possible to be represented by the $SE_2(3)$ group (a common treatment is augmenting the bias by another direct product group). Secondly, it supposes a flat earth surface and no earth rotation. The curved earth surface introduces position-related gravity change term and earth rotation introduces velocity-related Coriolis force term.

At present, there are fewer systematic studies on the analysis on the error propagation autonomy. In the counterpart of this paper, the $SE_2(3)$ group based navigation models under three commonly used frames (inertial frame, earth frame and world frame) were revisited. The theoretical analysis of anonymous property of $SE_2(3)$ group based EKF has been given and we identified two approximate autonomy properties. Then we proposed a novel construct method for the non-inertial frame navigation model by representing the velocity in the inertial frame to improve autonomy. This paper mainly illustrates the improved $SE_2(3)$ group based EKF algorithms by a Monte Carlo simulation and real-world SINS/ODO integrated navigation results.

## II. THE SYSTEM EQUATION AND MEASUREMENT EQUATION OF SINS/ODO INTEGRATED NAVIGATION

In the counterpart of this paper, the SINS/ODO integrated navigation model under different frames has been given. In this paper, the system equation and measurement equations of the world frame SINS/ODO integrated navigation further considering scale factor, lever arm and installation angle are given.

### A. The EKF's system and measurement equations

The system dynamic of EKF is modeled as in (1)
$$\dot{x}_{EKF} = F_{EKF} x_{EKF} + G_{EKF} w \quad (1)$$

The noise is modeled as in (2)
$$w = \begin{bmatrix} w_{gx} & w_{gy} & w_{gz} & w_{ax} & w_{ay} & w_{az} & w_{\delta k_{odo}} & w_{\alpha_\theta} \end{bmatrix}^T \quad (2)$$

where $w_{gi}$, $w_{ai} (i = x, y, z)$ stands for gyroscope and accelerometer measurement white noise and $w_{\delta k_{odo}}, w_{\alpha_\theta}$ stands for scale factor and installation angle white noise respectively.

The system state is defined as in (3)
$$x_{EKF} = \begin{bmatrix} (x_{EKF-NAV})^T & (x_{INS})^T & (x_{ODO})^T \end{bmatrix}^T$$
$$x_{EKF-NAV} = \begin{bmatrix} (\phi_{wb}^w)^T & (\delta v_{wb}^w)^T & (\delta r_{wb}^w)^T \end{bmatrix}^T \quad (3)$$
$$x_{INS} = \begin{bmatrix} (\varepsilon^b)^T & (\nabla^b)^T \end{bmatrix}^T, x_{ODO} = \begin{bmatrix} \delta k_{odo} & \alpha_\theta & \alpha_\psi & (\delta L^b)^T \end{bmatrix}^T$$

where the $x_{EKF-NAV}$, $x_{INS}$ and $x_{ODO}$ stands for navigation related state (attitude error $\phi_{wb}^w$, velocity error $\delta v_{wb}^w$ and position error $\delta r_{wb}^w$), inertial device related state (gyroscope constant bias $\varepsilon^b$ and accelerometer constant bias $\nabla^b$) and odometer related state (scale factor $\delta k_{odo}$, pitch installation angle $\alpha_\theta$, yaw installation angle $\alpha_\psi$ and lever arm $\delta L^b$) respectively.

The superscript and subscript of vectors α, β and γ represent different coordinate system or termed frame: α represents the reference frame, β the body frame, and γ the projection frame. In this paper, the reference frame and the projection frame are selected from the inertial frame (*i*-frame), the earth centered earth fixed frame (earth frame, *e*-frame) or the world frame (*w*-frame, an earth fixed tangent plane frame).

The corresponding system equations are in (4)-(6)

$$F_{EKF} = \begin{bmatrix} F_{EKF-NAV} & F_{EKF-INS} & 0_{9\times 6} \\ 0_{6\times 9} & 0_{6\times 6} & 0_{6\times 6} \\ 0_{6\times 9} & 0_{6\times 6} & F_{ODO} \end{bmatrix} \quad (4)$$

$$G_{EKF} = \begin{bmatrix} G_{EKF-INS} & 0_{6\times 2} \\ 0_{2\times 6} & G_{ODO} \end{bmatrix}$$

$$F_{EKF-NAV} = \begin{bmatrix} -(\omega_{ie}^w \times) & 0_{3\times 3} & 0_{3\times 3} \\ (\tilde{C}_b^w f_{ib}^b \times) & -2(\omega_{ie}^w \times) & 0_{3\times 3} \\ 0_{3\times 3} & I_3 & 0_{3\times 3} \end{bmatrix} \quad (5)$$

$$F_{EKF-INS} = \begin{bmatrix} -\tilde{C}_b^w & 0_{3\times 3} \\ 0_{3\times 3} & \tilde{C}_b^w \\ 0_{3\times 3} & 0_{3\times 3} \end{bmatrix}, G_{EKF-INS} = \begin{bmatrix} -\tilde{C}_b^w & 0_{3\times 3} \\ 0_{3\times 3} & \tilde{C}_b^w \\ 0_{3\times 3} & 0_{3\times 3} \end{bmatrix}$$

$$F_{ODO} = \begin{bmatrix} -1/\tau_{\delta k_{odo}} & 0 & 0 & 0_{1\times 3} \\ 0 & -1/\tau_{\alpha_\theta} & 0 & 0_{1\times 3} \\ 0 & 0 & -1/\tau_{\alpha_\psi} & 0_{1\times 3} \\ 0_{3\times 1} & 0_{3\times 1} & 0_{3\times 1} & 0_{3\times 3} \end{bmatrix}, G_{ODO} = \begin{bmatrix} w_{\delta k_{odo}} & 0 \\ 0 & w_{\alpha_\theta} \end{bmatrix} \quad (6)$$

where $\omega_{ie}^w$, $C_b^w$ and $f_{ib}^b$ stand for earth rotation vector, attitude direction cosine matrix (DCM) from body frame to world frame and specific force respectively. $0_{m\times n}$ is a m times n full zero matrix, $\times$ represents a skew-symmetric matrix and $\sim$ represents estimated variable.

The measurement equation is derived from the difference between the odometer velocity $v_{ODO}^m$ and velocity calculated by integrated navigation system $\tilde{v}_{wb}^m$. The superscript *m* stands for odometer frame and it satisfies $C_b^m \approx I + (\delta\alpha \times)$ where $\delta\alpha = [\alpha_\gamma \ \alpha_\theta \ \alpha_\psi]^T$ is installation angle vector. The roll installation $\alpha_\gamma$ is usually neglected since it has no effect on measurement.

The measurement equation of EKF is given in (7)

$$\begin{aligned} \delta z_v &= \tilde{v}_{wb}^m - v_{ODO}^m = \tilde{C}_w^m \tilde{v}_{wb}^w - \tilde{v}_{ODO}^m + v_v \\ &= C_b^m C_w^w (I + \phi_{wb}^w \times)(v_{wb}^w + \delta v_{wb}^w) - (1 + \delta k_D) v_{ODO}^m + v_v \\ &= -C_w^m (v_{wb}^w \times) \phi_{wb}^w + C_w^m \delta v_{wb}^w - v_D^m \delta k_{odo} \\ &\quad + (v_{ODO}^m \times)\delta\alpha - C_b^m (\omega_{wb}^b \times)\delta L^b + v_v \end{aligned} \quad (7)$$

where $v_v$ is measurement noise.

Rewrite (7) in a matrix form as in (8)

$$\delta z_v = H_{EKF} x_{EKF} + v_v \quad (8)$$

where

$$H_{EKF} = \begin{bmatrix} -C_w^m (v_{wb}^w \times) & C_w^m & 0_{3\times 9} & M_v^m & -C_b^m (\omega_{wb}^b \times) \end{bmatrix} \quad (9)$$

$$v_{ODO}^m = [v_D \ 0 \ 0]^T, M_v^m = \begin{bmatrix} -v_D & 0 & 0 \\ 0 & 0 & -v_D \\ 0 & v_D & 0 \end{bmatrix} \quad (10)$$

and $v_D$ is the front speed measured by odometer and $\omega_{wb}^b$ is the angular velocity in the body frame.

### B. The LG-EKF-R's system and measurement equations

The system dynamic of LG-EKF-R is modeled as:

$$\dot{x}_{LG-EKF-R} = F_{LG-EKF-R} x_{LG-EKF-R} + G_{LG-EKF-R} w \quad (11)$$

where the system state is defined as in (12) and the noise vector is the same as (2).

$$x_{LG-EKF-R} = \begin{bmatrix} (x_{LG-EKF-R-NAV})^T & (x_{INS})^T & (x_{ODO})^T \end{bmatrix}^T \quad (12)$$

$$x_{LG-EKF-R-NAV} = \begin{bmatrix} (\phi_{wb}^w)^T & (J\rho_{vwb}^w)^T & (J\rho_{rwb}^w)^T \end{bmatrix}^T$$

where

$$\begin{aligned} J\rho_{vwb}^w &\approx -\delta v_{wb}^w + v_{wb}^w \times \phi_{wb}^w \\ J\rho_{rwb}^w &\approx -\delta r_{wb}^w + (r_{wb}^w - r_{wb(0)}^w) \times \phi_{wb}^w \end{aligned} \quad (13)$$

The system equation of LG-EKF-R is given as in (14)

$$F_{LG-EKF-R} = \begin{bmatrix} F_{LG-EKF-R-NAV} & F_{LG-EKF-R-INS} & 0_{9\times 6} \\ 0_{6\times 9} & 0_{6\times 6} & 0_{6\times 6} \\ 0_{6\times 9} & 0_{6\times 6} & F_{ODO} \end{bmatrix} \quad (14)$$

$$G_{LG-EKF-R} = \begin{bmatrix} G_{LG-EKF-R-INS} & 0_{6\times 2} \\ 0_{2\times 6} & G_{ODO} \end{bmatrix}$$

where

$$F_{LG-EKF-R-NAV} = \begin{bmatrix} -(\tilde{\omega}_{ib}^w \times) & 0_{3\times 3} & 0_{3\times 3} \\ (g_{wb}^w + (v_{wb}^w \times)(\omega_{ie}^w \times)) & -(2\omega_{ie}^w \times) & 0_{3\times 3} \\ -(\tilde{r}_{wb}^w - r_{wb(0)}^w \times)(\omega_{ie}^w \times) & I_3 & 0_{3\times 3} \end{bmatrix}$$

$$F_{LG-EKF-R-INS} = \begin{bmatrix} -\tilde{C}_b^w & 0_{3\times 3} \\ -(v_{wb}^w \times) C_b^w & -\tilde{C}_b^w \\ -(\tilde{r}_{wb}^w - r_{wb(0)}^w \times) C_b^w & 0_{3\times 3} \end{bmatrix} \quad (15)$$

$$G_{LG-EKF-R-INS} = \begin{bmatrix} -\tilde{C}_b^w & 0_{3\times 3} \\ -(v_{wb}^w \times) C_b^w & -\tilde{C}_b^w \\ -(\tilde{r}_{wb}^w - r_{wb(0)}^w \times) C_b^w & 0_{3\times 3} \end{bmatrix}$$

where $\omega_{ib}^w$, $g_{wb}^w$ stand for world frame rotation velocity vector measured by gyroscope and gravity vector respectively. $v_{wb}^w$, $r_{wb}^w$ and $r_{wb(0)}^w$ is the world frame velocity, position and initial position vector respectively.

The measurement equation of LG-EKF-R can be derived from (7) and given in (16)

$$\delta z_v = -C_w^m J\rho_{vwb}^w - v_D^m \delta k_{odo} + (v_{ODO}^m \times)\delta\alpha - C_b^m (\omega_{wb}^b \times)\delta L^b + v_v \quad (16)$$

where the approximation of $J\rho_{vwb}^w \approx -\delta v_{wb}^w + \delta v_{wb}^w \times \phi_{wb}^w$ is used.

Rewrite (16) in a matrix form as in (17)

$$\delta z_v = H_{LG-EKF-R} x_{LG-EKF-R} + v_v \quad (17)$$

$$H_{LG-EKF-R} = \begin{bmatrix} 0_{3\times3} & -C_w^m & 0_{3\times9} & M_v^m & -C_b^m(\omega_{wb}^b \times) \end{bmatrix} \quad (18)$$

### C. The LG-EKF-L's system and measurement equations

The system dynamic of LG-EKF-L is modeled as:

$$\dot{x}_{LG-EKF-L} = F_{LG-EKF-L} x_{LG-EKF-L} + G_{LG-EKF-L} w \quad (19)$$

where the system state is defined as in (20) and the noise vector is the same as (2).

$$x_{LG-EKF-L} = \begin{bmatrix} (x_{LG-EKF-L-NAV})^T & (x_{INS})^T & (x_{ODO})^T \end{bmatrix}^T$$
$$x_{LG-EKF-L-NAV} = \begin{bmatrix} (\phi_{wb}^b)^T & (-\delta v_{wb}^b)^T & (-\delta r_{wb}^b)^T \end{bmatrix}^T \quad (20)$$

The system equation of LG-EKF-L is given as in (21)

$$F_{LG-EKF-L} = \begin{bmatrix} F_{LG-EKF-L-NAV} & F_{LG-EKF-L-INS} & 0_{9\times6} \\ 0_{6\times9} & 0_{6\times6} & 0_{6\times6} \\ 0_{6\times9} & 0_{6\times6} & F_{ODO} \end{bmatrix} \quad (21)$$

$$G_{LG-EKF-L} = \begin{bmatrix} G_{LG-EKF-L-INS} & 0_{6\times2} \\ 0_{2\times6} & G_{ODO} \end{bmatrix}$$

$$F_{LG-EKF-L-NAV} = \begin{bmatrix} -(\tilde{\omega}_{ib}^b \times) & 0_{3\times3} & 0_{3\times3} \\ (\tilde{f}_{ib}^b \times) & -(\omega_{ib}^b + \omega_{ie}^b) & 0_{3\times3} \\ 0_{3\times3} & I_3 & -(\omega_{ib}^b - \omega_{ie}^b) \end{bmatrix} \quad (22)$$

$$F_{LG-EKF-L-INS} = \begin{bmatrix} I_3 & 0_{3\times3} \\ 0_{3\times3} & -I_3 \\ 0_{3\times3} & 0_{3\times3} \end{bmatrix}, G_{LG-EKF-L-INS} = \begin{bmatrix} I_3 & 0_{3\times3} \\ 0_{3\times3} & -I_3 \\ 0_{3\times3} & 0_{3\times3} \end{bmatrix}$$

where $\omega_{ib}^b$ stands for body frame rotation velocity vector measured by gyroscope.

The measurement equation of LG-EKF-L is given in (23)

$$\delta z_v = \tilde{v}_{wb}^m - v_{ODO}^m = \tilde{C}_w^m \tilde{v}_{wb}^w - \tilde{v}_{ODO}^m + v_v$$
$$= C_b^m (I - \phi_{wb}^b \times) C_w^b (v_{wb}^w + \delta v_{wb}^w) - (1 + \delta k_{odo}) v_{ODO}^m + v_v$$
$$\approx (v_{wb}^m \times) C_b^m \phi_{wb}^b + \delta v_{wb}^m - v_D^m \delta k_{odo} + (v_{ODO}^m \times) \delta \alpha \quad (23)$$
$$- C_b^m (\omega_{wb}^b \times) \delta L^b + v_v$$

Rewrite (23) in a matrix form as in (24)

$$\delta z_v = H_{LG-EKF-L} x_{LG-EKF-L} + v_v \quad (24)$$

$$H_{LG-EKF-L} = \begin{bmatrix} (v_{wb}^m \times) C_b^m & -C_b^m & 0_{3\times9} & M_v^m & -C_b^m(\omega_{wb}^b \times) \end{bmatrix} \quad (25)$$

### D. The A-LG-EKF-R's system and measurement equations

The system dynamic of the proposed A-LG-EKF-R is modeled as:

$$\dot{x}_{A-LG-EKF-R} = F_{A-LG-EKF-R} x_{A-LG-EKF-R} + G_{A-LG-EKF-R} w \quad (26)$$

where the system state is defined as (27) and the noise vector is the same as (2)

$$x_{A-LG-EKF-R} = \begin{bmatrix} (x_{A-LG-EKF-R-NAV})^T & (x_{INS})^T & (x_{ODO})^T \end{bmatrix}^T$$
$$x_{A-LG-EKF-R-NAV} = \begin{bmatrix} (\phi_{wb}^w)^T & (J\rho_{vib}^w)^T & (J\rho_{rwb}^w)^T \end{bmatrix}^T \quad (27)$$

where

$$J\rho_{vib}^{wR} \approx -\delta v_{ib}^w + \left[ (\tilde{v}_{ib}^w - (\omega_{ie}^w \times)(r_{ew}^w + r_{wb(0)}^w)) \times \right] \phi_{wb}^w$$
$$J\rho_{rwb}^{wR} \approx -\delta r_{wb}^w + \left[ (\tilde{r}_{wb}^w - r_{wb(0)}^w) \times \right] \phi_{wb}^w \quad (28)$$

The system equation is given in (29)

$$F_{A-LG-EKF-R} = \begin{bmatrix} F_{A-LG-EKF-R-NAV} & F_{A-LG-EKF-R-INS} & 0_{9\times6} \\ 0_{6\times9} & 0_{6\times6} & 0_{6\times6} \\ 0_{6\times9} & 0_{6\times6} & F_{ODO} \end{bmatrix} \quad (29)$$

$$G_{A-LG-EKF-R} = \begin{bmatrix} G_{A-LG-EKF-R-INS} & 0_{6\times2} \\ 0_{2\times6} & G_{ODO} \end{bmatrix}$$

where

$$F_{A-LG-EKF-R-NAV} = \begin{bmatrix} -(\omega_{ie}^w \times) & 0_{3\times3} & 0_{3\times3} \\ (\gamma_{ib}^w - (\omega_{ie}^w \times)^2 (r_{ew}^w + r_{wb(0)}^w)) \times & -(\omega_{ie}^w \times) & 0_{3\times3} \\ 0_{3\times3} & I_3 & -(\omega_{ie}^w \times) \end{bmatrix}$$

$$F_{A-LG-EKF-R-INS} = \begin{bmatrix} -\tilde{C}_b^w & 0_{3\times3} \\ -\delta\gamma_{ib}^w - \left[ (\tilde{v}_{ib}^w - (\omega_{ie}^w \times)(r_{ew}^w + r_{wb(0)}^w)) \times \right] \tilde{C}_b^w & -\tilde{C}_b^w \\ -(\tilde{r}_{wb}^w - r_{wb(0)}^w \times) C_b^w & 0_{3\times3} \end{bmatrix} \quad (30)$$

$$G_{A-LG-EKF-R-INS} = \begin{bmatrix} -\tilde{C}_b^w & 0_{3\times3} \\ -\delta\gamma_{ib}^w - \left[ (\tilde{v}_{ib}^w - (\omega_{ie}^w \times)(r_{ew}^w + r_{wb(0)}^w)) \times \right] \tilde{C}_b^w & -\tilde{C}_b^w \\ -(\tilde{r}_{wb}^w - r_{wb(0)}^w \times) C_b^w & 0_{3\times3} \end{bmatrix}$$

The measurement equation of A-LG-EKF-R can be derived as (31)

$$\delta z_v = -C_w^m J\rho_{vib}^{wR} + C_w^m (\omega_{ie}^w \times) J\rho_{rib}^{wR} - v_D^m \delta k_{odo}$$
$$+ (v_{ODO}^m \times) \delta \alpha - C_b^m(\omega_{wb}^b \times) \delta L^b + v_v \quad (31)$$

Rewrite (31) in a matrix form as in (32)

$$\delta z_v = H_{A-LG-EKF-R} x_{A-LG-EKF-R} + v_v \quad (32)$$

$$H_{A-LG-EKF-R} = \begin{bmatrix} 0_{3\times3} & -C_w^m & C_w^m(\omega_{ie}^w \times) & 0_{3\times6} & M_v^m & -C_b^m(\omega_{wb}^b \times) \end{bmatrix} \quad (33)$$

### E. The A-LG-EKF-L's system and measurement equations

The system dynamic of the proposed A-LG-EKF-L is modeled as in (34)

$$\dot{x}_{A-LG-EKF-L} = F_{A-LG-EKF-L} x_{A-LG-EKF-L} + G_{A-LG-EKF-L} w \quad (34)$$

where the system state is defined as in (35) and the noise vector is the same as (2).

$$x_{A-LG-EKF-L} = \left[ (x_{A-LG-EKF-L-NAV})^T \ (x_{INS})^T \ (x_{ODO})^T \right]^T$$
$$x_{A-LG-EKF-L-NAV} = \left[ (\phi_{wb}^b)^T \ (-\delta v_{ib}^b)^T \ (-\delta r_{wb}^b)^T \right]^T \quad (35)$$

The system equation of A-LG-EKF-L is given in (36)

$$F_{A-LG-EKF-L} = \begin{bmatrix} F_{A-LG-EKF-L-NAV} & F_{A-LG-EKF-L-INS} & 0_{9\times6} \\ 0_{6\times9} & 0_{6\times6} & 0_{6\times6} \\ 0_{6\times9} & 0_{6\times6} & F_{ODO} \end{bmatrix} \quad (36)$$

$$G_{A-LG-EKF-L-INS} = \begin{bmatrix} G_{A-LG-EKF-L-INS} & 0_{6\times2} \\ 0_{2\times6} & G_{ODO} \end{bmatrix}$$

where

$$F_{A-LG-EKF-L-NAV} = \begin{bmatrix} -(\tilde{\omega}_{ib}^b \times) & 0_{3\times3} & 0_{3\times3} \\ (\tilde{f}_{ib}^b \times) & -(\tilde{\omega}_{ib}^b \times) & 0_{3\times3} \\ 0_{3\times3} & I_3 & -(\tilde{\omega}_{ib}^b \times) \end{bmatrix} \quad (37)$$

$$F_{A-LG-EKF-L-INS} = \begin{bmatrix} I_3 & 0_{3\times3} \\ 0_{3\times3} & -I_3 \\ 0_{3\times3} & 0_{3\times3} \end{bmatrix}, G_{A-LG-EKF-L-INS} = \begin{bmatrix} I_3 & 0_{3\times3} \\ 0_{3\times3} & -I_3 \\ 0_{3\times3} & 0_{3\times3} \end{bmatrix}$$

The measurement equation is derived as in (38)

$$\delta z_v = \tilde{v}_{wb}^m - v_{ODO}^m$$
$$= (v_{wb}^m \times) C_b^m \phi_{wb}^b + C_b^m \delta v_{ib}^b - C_b^m (\omega_{ie}^b \times) \delta r_{wb}^b \quad (38)$$
$$- v_D^m \delta k_{odo} + (v_{ODO}^m \times) \delta \alpha - C_b^m (\omega_{wb}^b \times) \delta L^b + v_v$$

Rewrite (38) in a matrix form as in (39)

$$\delta z_v = H_{A-LG-EKF-L} x_{A-LG-EKF-L} + v_v \quad (39)$$

$$H_{A-LG-EKF-L} = \left[ (v_{wb}^m \times) C_b^m \ -C_b^m \ C_b^m (\omega_{ie}^b \times) \ 0_{3\times6} \ M_v^m \ -C_b^m (\omega_{wb}^b \times) \right] \quad (40)$$

## III. MONTE-CARLO SIMULATION

To evaluate the performance of the filter using a more anomaly property, a Monte-Carlo simulation of ground vehicle is provided in this section. A ground vehicle is modeled, and it is equipped with strapdown inertial navigation system (SINS) and odometer (ODO) for integrated navigation. The evaluated algorithms are as in TABLE I.

The propagation rate and update rate are closely related to the inertial device and odometer measurement respectively. The system matrix of right form $SE_2(3)$ group based navigation models as shown in (15) and (30) is formed by the slow changing earth rotation and earth gravity terms. It guarantees enough accuracy of prediction process of Kalman filter even at lower propagation rate. On the contrary, the left form $SE_2(3)$ group based navigation models as shown in (22) and (37) have to propagate as fast as possible to capture the fast changing specific force. This difference makes right error form navigation models more computationally efficient. In this paper, the propagation and update rate factor is not discussed, and they are set as 100Hz and 1Hz respectively in all algorithms.

TABLE I
Algorithms to be evaluated

| Name | Reference frame | Group | Error form |
|---|---|---|---|
| SO-EKF-*i* | Inertial frame | SO3×R6×R6 | Linear error |
| LSE-EKF-*i* | | SE23×R6×R6 | Left nonlinear error |
| RSE-EKF-*i* | | SE23×R6×R6 | Right nonlinear error |
| SO-EKF-*e* | Earth frame | SO3×R6×R6 | Linear error |
| LSE-EKF-*e* | | SE23×R6×R6 | Left nonlinear error |
| A-LSE-EKF-*e* | | SE23×R6×R6 | Proposed left error |
| RSE-EKF-*e* | | SE23×R6×R6 | Right nonlinear error |
| A-RSE-EKF-*e* | | SE23×R6×R6 | Proposed right error |
| SO-EKF-*w* | World frame | SO3×R6×R6 | Linear error |
| LSE-EKF-*w* | | SE23×R6×R6 | Left nonlinear error |
| A-LSE-EKF-*w* | | SE23×R6×R6 | Proposed left error |
| RSE-EKF-*w* | | SE23×R6×R6 | Right nonlinear error |
| A-RSE-EKF-*w* | | SE23×R6×R6 | Proposed right error |

### A. Simulation Data Generation

To simulate the trajectory with better fidelity, it is composed of distinct phases including straight-line motion, turning maneuvers, and ascending segments. It starts with a stationary period of 100 seconds at the initial position (marked in green in Fig 1), then moves forward with an acceleration of 0.1 m/s² for 100 seconds and moves forward for another 200 seconds. After that, it turns left at an angular velocity of 2 deg/s for 225 seconds, covering a total of 450 degrees. It then moves forward for another 200 seconds, turns right at the same angular velocity of 2 deg/s for 225 seconds, and moves forward for another 200 seconds. It then climbs slowly, ascending approximately 3.5 meters within 10 seconds. It moves forward for another 200 seconds, repeats the same left and right turns, and finally decelerates to an ending point (marked in red in Fig 1). The total duration is 2510 seconds, and the total distance traveled is approximately 22 kilometers.

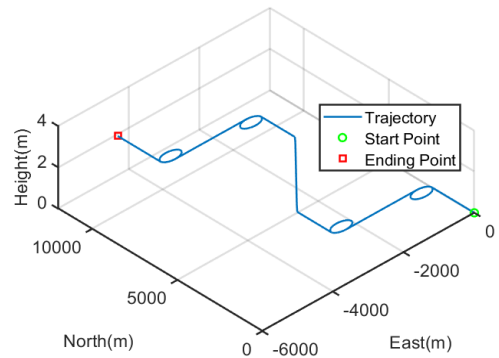

Fig. 1. The trajectory of the ground vehicle.

In this simulation, the odometer (ODO) and strap-down inertial navigation system (SINS) are modelled. The inertial navigation system consists of three gyroscopes and three accelerometers with 200Hz output frequency. The odometer has 1Hz output frequency, and its velocity noise strength is set as 0.1 m/s. The initial attitude and constant inertial sensor bias are set randomly as shown in TABLE II. The odometer is assumed the scale factor and installation angle as well as lever arm are calibrated very well before experiment and be simply neglected in the simulation.

TABLE II
**Monte-Carlo Simulation Setting**

| Parameter | | Distribution |
|---|---|---|
| gyroscope constant bias (°/h) | | U(−0.1,0.1) |
| accelerometer constant bias (μg) | | U(−100,100) |
| angular random walk (°/sqrt(h)) | | Constant(0.005) |
| velocity random walk (m/s/sqrt(h)) | | Constant(1.2×10⁻²) |
| Initial attitude | Roll (°) | U(−1,1) |
| | Pitch (°) | U(−1,1) |
| | Yaw (°) | U(−30,30) |

In order to simulate the impact of engine vibration of the vehicle, random vibrations are added onto inertial device output. The vibration is a sinusoidal wave noise with a random frequency of 333Hz, but its amplitude is modulated by a random signal. The amplitude modulation signal has an amplitude less than 0.1g.

**B. Results and Discussion**

The attitude, velocity and position error are the arithmetic mean values of the results obtained from $M$ times of the Monte Carlo simulation. They are shown in (41) and $M$ is set to be 200 here.

$$E_{att} = \frac{1}{M}\sum_{i=1}^{M}\left(\hat{\boldsymbol{\theta}}_{i,k} - \boldsymbol{\theta}_{i,k}\right)$$
$$E_{vel} = \frac{1}{M}\sum_{i=1}^{M}\left(\hat{\boldsymbol{v}}_{i,k} - \boldsymbol{v}_{i,k}\right) \quad (41)$$
$$E_{Hpos} = \frac{1}{M}\sum_{i=1}^{M}\left\|\hat{\boldsymbol{r}}_{i,k} - \boldsymbol{r}_{i,k}\right\|^2$$

where $k$ represents the time step and superscript $\hat{\cdot}$ stands for calculated navigation variable.

**1) Inertial frame filtering algorithms**

The attitude error, velocity error and position error of algorithms under inertial frame are shown in Fig. 2 - Fig. 4.

As can be seen from Fig. 2 that the LG-EKF-i-R achieves lower average yaw error compared to LG-EKF-i-L as well as EKF-i, which is benefited from the nonlinear nature of right error form of Lie group navigation state.

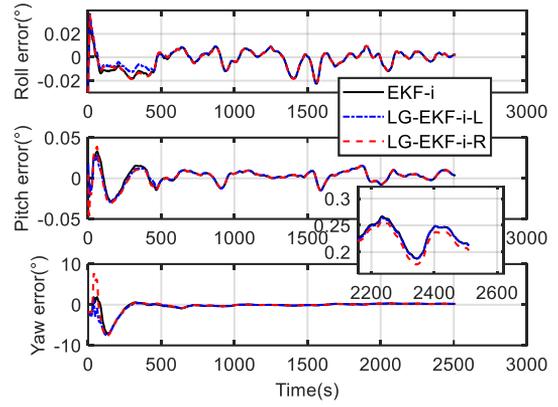

Fig. 2. The attitude error of inertial frame algorithms.

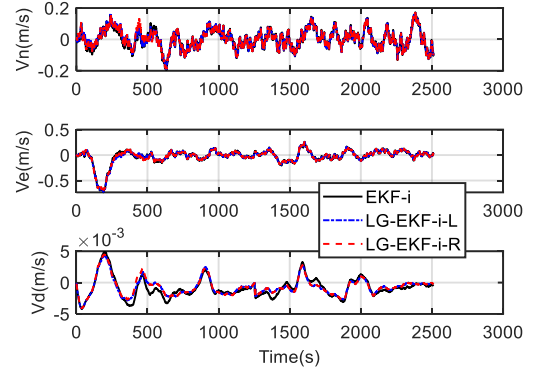

Fig. 3. The velocity error of inertial frame algorithms.

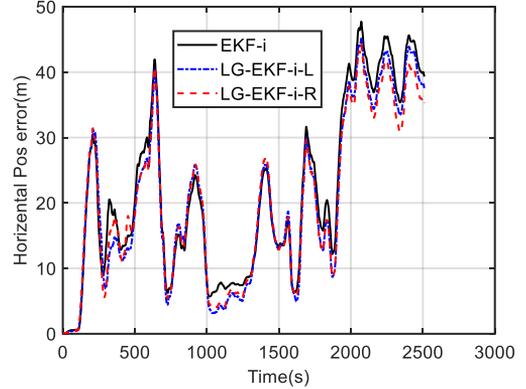

Fig. 4. The position error of inertial frame algorithms.

The velocity errors of the three algorithms are nearly the same since the odometer provides fully observable velocity. Fig. 4 shows LG-EKF-i-R achieves lowest positioning error, which benefited from that the body velocity measurement is of right invariant type [21].

**2) Earth frame filtering algorithms**

The attitude error, velocity error and position error of algorithms under earth frame are shown in Fig. 5 to Fig. 7.

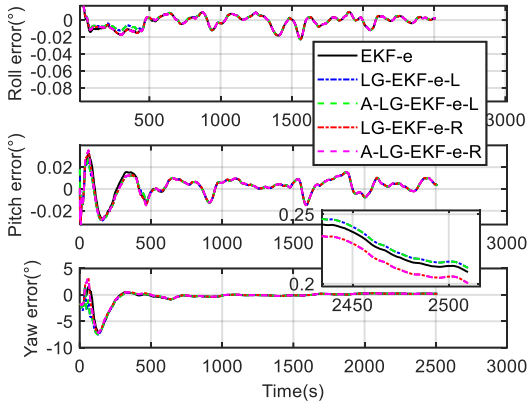

Fig. 5. The attitude error of earth frame algorithms.

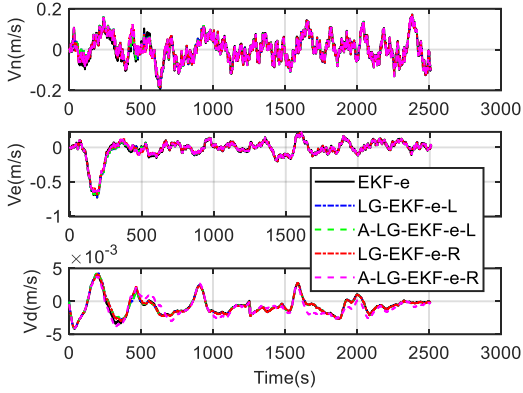

Fig. 6. The velocity error of earth frame algorithms.

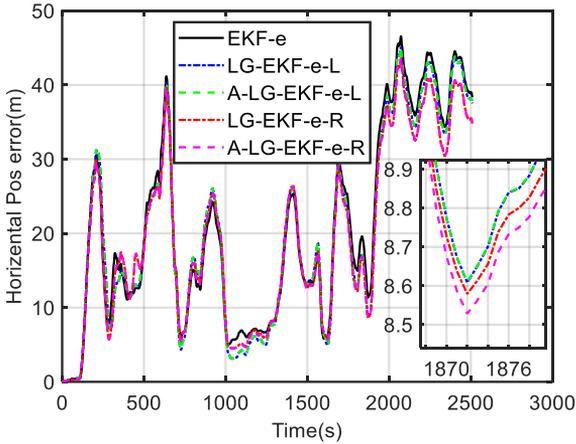

Fig. 7. The position error of earth frame algorithms.

As can be seen from Fig. 5 and Fig. 7, the LG-EKF-e-R achieves lower attitude and positioning error. The improvement of the proposed left error form A-LG-EKF-e-L compared to LG-EKF-e-L is not obvious.

3) **World frame filtering algorithms**

The attitude error, velocity error and position error results of algorithms under world frame are shown in Fig. 8 to Fig. 10.

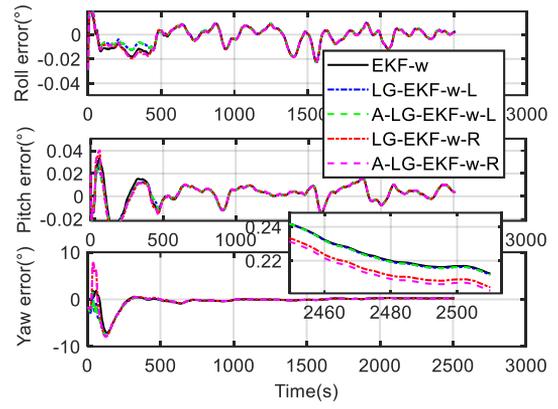

Fig. 8. The attitude error of world frame algorithms.

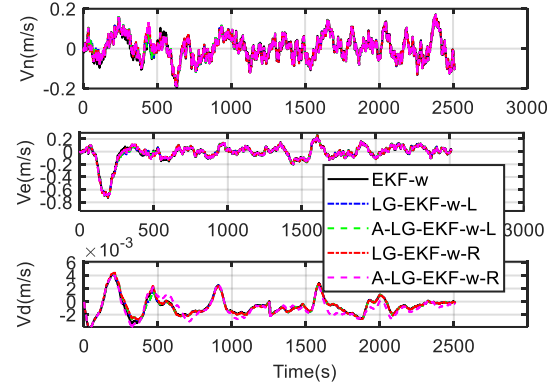

Fig. 9. The velocity error of world frame algorithms.

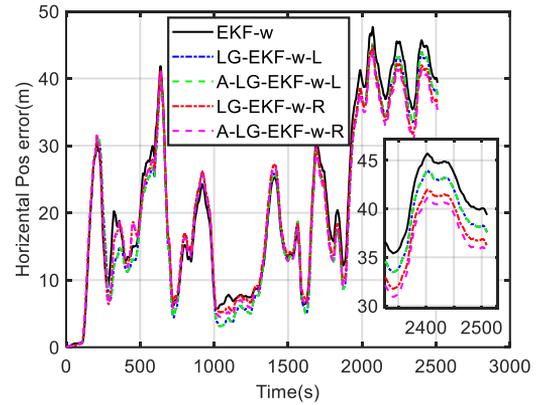

Fig. 10. The position error of world frame algorithms.

As can be seen from Fig. 8 and Fig. 10, the LG-EKF-w-R achieves lowest attitude and positioning error.

The statistical positioning error results of navigation algorithm in different frames are shown in Table III. It can be seen that the proposed algorithms (A-R/LSE-EKFs) obtain some lower positioning error compared to the original SE2(3) group based algorithms (R/LSE-EKFs) in earth and world frames as well as SO(3) group based algorithms (SO-EKFs).

### TABLE III
### Comparison position error of different algorithms

| Algorithms | Reference frame | Horizontal Position RMSE (m) |
|---|---|---|
| SO-EKF-*i* | Inertial frame | 25.32 |
| LSE-EKF-*i* | | 24.01 |
| RSE-EKF-*i* | | **23.46** |
| SO-EKF-*e* | Earth frame | 24.55 |
| LSE-EKF-*e* | | 23.98 |
| A-LSE-EKF-*e* (proposed) | | 23.98 |
| RSE-EKF-*e* | | 23.05 |
| A-RSE-EKF-*e* (proposed) | | **23.04** |
| SO-EKF-*w* | World frame | 25.31 |
| LSE-EKF-*w* | | 24.01 |
| A-LSE-EKF-*w* (proposed) | | 24.00 |
| RSE-EKF-*w* | | 24.11 |
| A-RSE-EKF-*w* (proposed) | | **23.58** |

## IV. REAL-WORLD EXPERIMENT

### A. Experiment description

In this section, the SINS/ODO integrated navigation experiment has been conducted. The experiment consists of 6 individual trajectories, and they are shown in Fig 11. The green circle and red square stand for beginning point and ending point respectively.

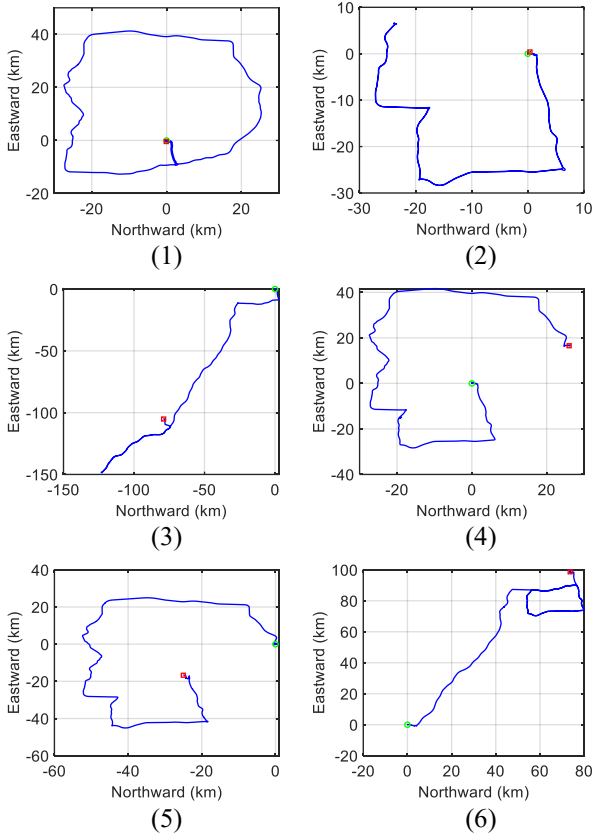

Fig. 11. The trajectories of the ground vehicle.

The ground truth of position comes from the GNSS/INS integrated navigation, which has an accuracy of nearly 1m. The IMU consists three high-precision gyroscopes and accelerometers with bias lower than 0.001 °/h and 20 μg respectively. The odometer has a velocity uncertainty of approximately 0.05m/s.

### B. Experiment result and discussion

The SINS/ODO integrated navigation results are compared in Fig. 12. For Due to the limitation of space, only the world frame algorithm is evaluated.

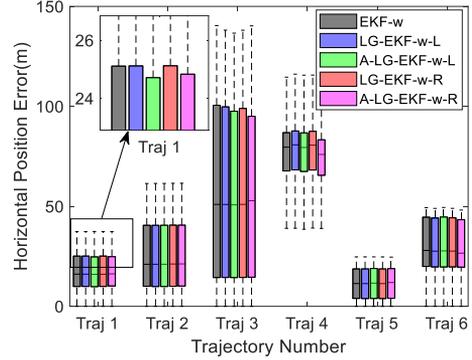

Fig. 12. The horizontal position error of SINS/ODO integrated navigation.

The SINS/ODO integrated navigation results are summarized in Table IV. The '-EKF-w' attached to the algorithm name is omitted in Table IV for conciseness.

### TABLE IV
### Positioning error of algorithms under world frame

| Trajectory number | Horizontal Position RMSE (m) | | | | |
|---|---|---|---|---|---|
| | SO | LSE | RSE | A-LSE | A-RSE |
| 1 | 31.6 | **31.5** | 31.6 | 31.6 | 31.6 |
| 2 | 20.2 | 20.2 | 20.2 | **20.1** | 20.2 |
| 3 | 73.5 | 72.6 | 72.5 | **71.6** | 72.4 |
| 4 | 13.6 | 13.5 | **13.5** | 13.7 | 13.7 |
| 5 | 76.3 | 77.6 | 77.4 | 76.6 | **73.8** |
| 6 | 34.3 | 33.9 | 34.0 | 34.3 | **33.5** |
| Average | 43.6 | 43.6 | 43.5 | 43.3 | **42.7** |

As can be seen from Fig. 12 and Table IV, the proposed A-RSE-EKF-*w* algorithm achieved the lowest average positioning accuracy in six independent experiments. The positioning accuracy improvement of the proposed algorithms mainly comes from the enhancement of the autonomy of original LG-EKFs.

## V. CONCLUSION

These two papers dug deeply into the autonomy of $SE_2(3)$ group based extended Kalman filter. In this paper, a Monte-Carlo simulation and a real-world SINS/ODO integrated navigation experiment are conducted to verify the theoretical analysis given in previous paper. By using the construction method to replace the non-inertial velocity with inertial velocity, the positioning error has been further decreased compared to the original $SE_2(3)$ group based navigation model. The results demonstrated performance improvement of $SE_2(3)$ group based navigation model with improved autonomy compared to the original ones. The conclusion could be further expanded to other autonomous navigation applications.

# REFERENCES

[1] R. E. Kalman, "A New Approach to Linear Filtering and Prediction Problems." *Trans. ASME*, *J. Basic Eng.*, vol. Series D 82, pp. 35-45, Mar. 1960.

[2] R. E. Kalman, and R.S. Bucy, "New Results in Linear Filtering and Prediction Theory," *Trans. ASME*, *J. Basic Eng.*, vol. Series D 82, pp. 95-108, Mar. 1961.

[3] S. F. Schmidt, "The Kalman Filter: Its Recognition and Development for Aerospace Applications," *J. Guid. Control*, vol. 4, pp. 4-7, Jan.-Feb. 1981.

[4] E. J. Lefferts, F. L. Markley, and M. D. Shuster, "Kalman Filtering for Spacecraft Attitude Estimation," *J. Guid., Control, Dyn.*, vol. 5, no. 5, pp. 417–429, Sep. 1982.

[5] V. J. Aidala, "Kalman Filter Behavior in Bearings-Only Tracking Applications," *IEEE Trans. Aerosp. Electron. Syst.*, vol. AES-15, no. 1, pp. 29-39, Jan. 1979.

[6] Y. Huang, Y. Zhang, N. Li, Z. Wu and J. A. Chambers. "A novel robust student's t-based Kalman filter." *IEEE Trans. Aerosp. Electron. Syst.*, vol. 53, no. 3, pp. 1545-1554, June 2017.

[7] M. S. Grewal and A. P. Andrews, "Applications of Kalman Filtering in Aerospace 1960 to the Present," *IEEE Control Syst. Mag.*, vol. 30, no. 3, pp. 69-78, June 2010.

[8] F. L. Markley, "Attitude Error Representations for Kalman Filtering," *J. Guid., Control, Dyn.*, vol. 26, no. 2, pp. 311–317, Mar. 2003.

[9] H. A. Hashim, L. J. Brown and K. McIsaac, "Nonlinear Pose Filters on the Special Euclidean Group SE(3) With Guaranteed Transient and Steady-State Performance," *IEEE Trans. Syst., Man, Cybern.*, vol. 51, no. 5, pp. 2949-2962, May 2021.

[10] T. Qin, P. Li and S. Shen, "VINS-Mono: A Robust and Versatile Monocular Visual-Inertial State Estimator," *IEEE Trans. Robot.*, vol. 34, no. 4, pp. 1004-1020, Aug. 2018.

[11] M. Zhang, X. Zuo, Y. Chen, Y. Liu and M. Li, "Pose Estimation for Ground Robots: On Manifold Representation, Integration, Reparameterization, and Optimization," *IEEE Trans. Robot.*, vol. 37, no. 4, pp. 1081-1099, Aug. 2021.

[12] A. Manzanilla, S. Reyes, M. Garcia, D. Mercado and R. Lozano, "Autonomous Navigation for Unmanned Underwater Vehicles: Real-Time Experiments Using Computer Vision," *IEEE Robot. Autom. Lett.*, vol. 4, no. 2, pp. 1351-1356, Apr. 2019.

[13] J. L. Crassidis. and F. L. Markley, "Unscented Filtering for Spacecraft Attitude Estimation," *J. Guid., Control, Dyn.*, vol. 26, no. 4, pp. 536–542, May 2003.

[14] F. Gustafsson et al., "Particle filters for positioning, navigation, and tracking," *IEEE Trans. Signal Process.*, vol. 50, no. 2, pp. 425-437, Feb. 2002.

[15] A. Barrau. "Non-linear state error based extended Kalman filters with applications to navigation," Ph.D. dissertation, Mines Paris Tech, Paris, France, 2015.

[16] A. Barrau and S. Bonnabel, "The Invariant Extended Kalman Filter as a Stable Observer," *IEEE Trans. Autom. Control*, vol. 62, no. 4, pp. 1797-1812, Apr. 2017.

[17] A. Barrau and S. Bonnabel, "The Geometry of Navigation Problems," *IEEE Trans. Autom. Control*, vol. 68, no. 2, pp. 689-704, Feb. 2023.

[18] M. Wang, J. Cui and W. Wu, "Left/Right Invariant Lie Group Error for SINS/GNSS Tightly Coupled Vehicular Integrated Navigation," *IEEE Trans. Veh. Technol.*, vol. 74, no. 6, pp. 8975-8988, June 2025.

[19] J. Cui, M. Wang, W. Wu and X. He. "Lie group based nonlinear state errors for MEMS-IMU/GNSS/magnetometer integrated navigation," *J. Navig.*, vol. 74, no. 4, pp. 887-900, Mar. 2021.

[20] J. Cui, M. Wang, W. Wu, R. Liu and W. Yang, "Enhanced LG-EKF Backtracking Framework for Body-Velocity-Aided Vehicular Integrated Navigation," *IEEE Trans. Veh. Technol.*, vol. 73, no. 10, pp. 14212-14223, Oct. 2024.

[21] L. Chang, H. Tang, G. Hu and J. Xu, "SINS/DVL Linear Initial Alignment Based on Lie Group SE3(3)," *IEEE Trans. Aerosp. Electron. Syst.*, vol. 59, no. 5, pp. 7203-7217, Oct. 2023.

[22] G. Hu, J. Geng, L. Chang, B. Gao and Y. Zhong, "Tightly Coupled SINS/DVL Based on Lie Group SE2(3) in Local-Level Frame," *IEEE Trans. Aerosp. Electron. Syst.*, [Early Access]. Available: https://ieeexplore.ieee.org/abstract/document/11098863, doi: 10.1109/TAES.2025.3593458.

[23] A. Fornasier, Y. Ng, R. Mahony and S. Weiss, "Equivariant Filter Design for Inertial Navigation Systems with Input Measurement Biases," 2022 *Proc. Int. Conf. Robot. Autom. (ICRA)*, Philadelphia, PA, USA, 2022, pp. 4333-4339.

[24] P. van Goor and R. Mahony, "EqVIO: An Equivariant Filter for Visual-Inertial Odometry," *IEEE Trans. Robot.*, vol. 39, no. 5, pp. 3567-3585, Oct. 2023.

[25] Y. Luo, C. Guo, & J. Liu, "Equivariant filtering framework for inertial-integrated navigation." *Satell. Navig.*, vol. 2, no. 30, pp. 1-17, Dec. 2021.

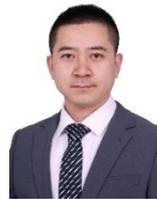

**Maosong Wang** received the B.S. degree from Harbin Engineering University in 2012, the M.S. degree from the National University of Defense Technology in 2014, and the Ph.D. degree from the National University of Defense Technology in 2018. From September 2016 to March 2018, he was a Visiting Student Researcher at the University of Calgary, Canada. Currently, he is an associated professor at the National University of Defense Technology. His research interests include inertial navigation algorithm, and multi-sensor integrated navigation theory and application

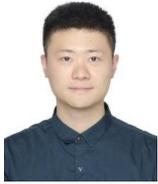

**Jiarui. Cui** received the B.S. degree from the Nanjing University of Aeronautics and Astronautics in 2018, the M.S. degree and Ph.D. degree from the National University of Defense Technology in 2020 and 2025. He is now an assistant researcher at Beijing Institute of Tracking and Telecommunications Technology. His research interests include inertial based multi-sensor integrated navigation and advanced state estimation algorithms.

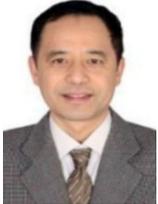

**Wenqi Wu** received the B.S. degree from Tianjin University in 1988, the M.S. degree from the National University of Defense Technology in 1991, and the Ph.D. degree from the National University of Defense Technology in 2002. Currently, he is a professor at the National University of Defense Technology. His research interests include GNSS, inertial, and multi-sensor integrated navigation theory and application.

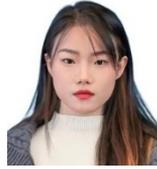

**Peiqi Li** received the B.S. degree from Harbin Engineering University in 2024.Currently,she is pursuing the degree of master in the National University of Defense Technology. Her research interests include inertial navigation algorithm, and Lie group based-EKF theory and application.

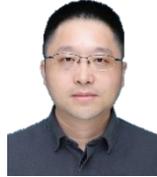

**Xianfei Pan** received the Ph.D. degree in control science and engineering from the National University of Defense Technology, Changsha, China, in 2008. Currently, he is a Professor of the College of Intelligence Science and Technology, National University of Defense Technology. His current research interests include multi-source integrated navigation, cooperative navigation and intelligent navigation.